\documentclass[journal]{IEEEtran}

\usepackage{cite}
\usepackage{graphicx}
\usepackage{amsmath,amssymb} 
\usepackage{enumerate}
\usepackage{wrapfig}
\usepackage{multicol}  
\usepackage{multirow}
\usepackage{subfigure}
\usepackage{caption}
\usepackage{booktabs} 
\usepackage{makecell}
\usepackage{url}

\usepackage[ruled]{algorithm2e}

\begin{document}

\title{Touchless Palmprint Recognition based on 3D Gabor Template and Block Feature Refinement}
\author{Zhaoqun~Li,
\and Xu~Liang,
\and Dandan~Fan,
\and Jinxing~Li,
\and Wei~Jia,~\IEEEmembership{Member,~IEEE},
\and David~Zhang*,~\IEEEmembership{Life~Fellow,~IEEE}

\thanks{Z. Li is with the School of Data Science, 
The Chinese University of Hong Kong, Shenzhen, 
and the Shenzhen Institute of Artificial Intelligence and Robotics for Society, 
Shenzhen 518172, China (e-mail: zhaoqunli@link.cuhk.edu.cn)}%
\thanks{X. Liang is with the Bio-Computing Research Center, 
School of Computer Science and Technology, 
Harbin Institute of Technology, Shenzhen, Shenzhen, 518055, China (e-mail: xuliangcs@gmail.com).}%
\thanks{D. Fan is with the School of Data Science, The Chinese University of Hong Kong, Shenzhen, 
and the Shenzhen Institute of Artificial Intelligence and Robotics for Society, Shenzhen 518172, China (e-mail: fanfan\_abu@163.com).}%
\thanks{J. Li is with the Bio-Computing Research Center, 
School of Computer Science and Technology, 
Harbin Institute of Technology, Shenzhen, Shenzhen, 518055, China (e-mail: lijinxing158@gmail.com).}%
\thanks{W. Jia is with the School of Computer and Information, Hefei University
of Technology, Hefei 230009, China (e-mail: china.jiawei@139.com).}%
\thanks{D. Zhang is with School of Data Science, Chinese University of Hong Kong (Shenzhen), 
and the Shenzhen Institute of Artificial Intelligence and Robotics for Society, Shenzhen 518172, China (e-mail:davidzhang@cuhk.edu.cn).}%
}

\markboth{Journal of \LaTeX\ Class Files,~Vol.~14, No.~8, August~2015}%
{Shell \MakeLowercase{\textit{et al.}}: Bare Demo of IEEEtran.cls for IEEE Journals}

\maketitle
 
\begin{abstract}
With the growing demand for hand hygiene and convenience of use, 
palmprint recognition with touchless manner made a great development recently,
providing an effective solution for person identification.
Despite many efforts that have been devoted to this area,
it is still uncertain about the discriminative ability of the touchless palmprint, 
especially for large-scale datasets.
To tackle the problem, in this paper,
we build a large-scale touchless palmprint dataset containing 2334 palms from 1167 individuals.
To our best knowledge, 
it is the largest touchless palmprint image benchmark ever collected with regard to the number of individuals and palms. 
Besides, 
we propose a novel deep learning framework for touchless palmprint recognition named 3DCPN (3D Convolution Palmprint recognition Network)
which leverages 3D convolution to dynamically integrate multiple Gabor features.
In 3DCPN, 
a novel variant of the Gabor filter is embedded into the first layer for enhancement of curve feature extraction.
With a well-designed ensemble scheme,
low-level 3D features are then convolved to extract high-level features.
Finally on the top, 
we set a region-based loss function to strengthen the discriminative ability of both global and local descriptors.
To demonstrate the superiority of our method, 
extensive experiments are conducted on our dataset and other popular databases TongJi and IITD, 
where the results show the proposed 3DCPN achieves state-of-the-art or comparable performances.
\end{abstract} 


\section{Introduction}
\label{sec_intro}
Biometric identification has been widely used in modern society, 
such as electronic payment, entrance control, and forensic identification.  
In the last decade,
we have witnessed the successful cases of biometric system using fingerprint~\cite{Cappelli2010Minutia,maltoni2009handbook}, 
iris~\cite{nalla2016toward,nguyen2017long} and face~\cite{Schroff2015facenet,Opitz2016grid,deng2019arcface}.
With the development of computer vision,
biometric based on image analysis become popular.
As a representative biometric technology, 
palmprint recognition provides a reliable and efficient solution for recognizing a person’s identity with high confidence ~\cite{Zhang2003Online,Liang2019A,Jia2012palmprint,Zhang2017Improving,Zhang2018Palmprint}.
Even in low resolution,
palmprint images contain rich biometric information and have high antispoof capability, 
which is desired for person identification~\cite{Zhang2003Online}. 
Based on the image capture manner,
palmprint images could be divided into two categories: touch-based and touchless.
In a real application, 
people prefer to use touchless manner which is more hygienic and convenient, 
especially under the current epidemic situation.
With this trend, 
a growing number of works turn to more challenged touchless palmprint recognition tasks.

Starting from PalmCode~\cite{Zhang2003Online},
a large number of works~\cite{jia2008palmprint,fei2016double,luo2016local,zhang2017towards} develop coding based methods.
In these methods, 
the orientation information or the texture information on the palm are encoded into a feature map.
With well-designed matching algorithms, 
coding based approaches are usually efficient.
However, the scheme of pixel-to-pixel comparison decreases their robustness. 
Recently more and more methods depending on machine learning techniques emerge.
Fei \emph{et al.}~\cite{fei2019learning} proposes a learned binary palmprint descriptor,
targeting to minimize the intra-class distance and maximize the inter-class distance.
Zhang \emph{et al.}~\cite{zhang2017towards} uses collaborative representation which learns the feature distribution in the training gallery, 
achieving high performances in identity recognition.
With the significant achievement of deep learning in computer vision, 
a lot of works~\cite{Dian2016Contactless,zhao2020deep,Shao_2019_CVPR_Workshops} employ convolutional neural network (CNN) as feature extractor.
Zhao \emph{et al.}~\cite{zhao2019joint} designs a network which is suitable for hyperspectral palmprint feature extraction.
In~\cite{matkowski2019palmprint},  
an end-to-end deep learning algorithm is proposed for accurate palmprint identification.
The framework has different parts which are responsible for hand image alignment and feature extraction respectively.
Genovese \emph{et al.}~\cite{Genovese2019palmnet} adopts adaptive Gabor filters and Principal Component Analysis (PCA) in a three-layer CNN 
that can learn discriminative features without class labels.

Though the research study of touchless palmprint recognition has been carried out for a long period,
the lack of large-scale palmprint datasets limits the further development of this domain to some extent.
Moreover, 
most existing methods are designed and verified on medium-scale datasets, 
such as PolyU~\cite{PolyU3DContactless}, TongJi~\cite{zhang2017towards} and IITD~\cite{IITD},
while the performance on a larger dataset remains unclear.
To address the above concerns, 
we build a large-scale touchless palmprint dataset that contains 2334 palms from 1167 individuals.
The embedded binocular camera in the acquisition device can capture palmprint image and palm vein image simultaneously.
In the dataset, 
each palm has 6 RGB images and 6 infra-red (IR) images with different postures and there are totally 28,008 images.
To the best of our knowledge, 
it is the largest touchless palmprint image benchmark ever collected with regard to the number of individuals and palms. 

We also notice that most existing CNN based methods leverage traditional neural architectures or simply embed handcrafted filters, 
while the potential of the Gabor filter is not fully exploited.
Based on our observation,
the lines on the palm are the most important parts in deep feature learning.
As a consequence,
a issue that degrades the recognition performance is the ROI misalignment.
In order to extract highly discriminative features,
we propose a novel deep learning framework named 3DCPN that takes full advantage of low-level Gabor features.
In the first layer of 3DCPN, 
multi-scale, multi-direction Gabor filters and curved Gabor filters are embedded.
Leveraging the steerability of Gabor filters, 
palmprint images are convolved with 3D Gabor templates that generate low-level features.
Then the features are reformatted to three spatial dimensions which involve width, height, and direction.
Several 3D convolution modules are designed for generating high-level palmprint descriptors.
Besides, 
a part of a palmprint is also a palmprint,
utilizing the local region information in palmprint recognition is an unavoidable topic.
In our method,
instead of simply extracting and fusing features in other methods~\cite{luo2016local},
we regard each subregion as a real palmprint.
Considering this local characteristic of palmprint,
we use a region-based loss function named block loss on the top of the network to enhance the block features.
The region-based learning scheme avoids the ROI misalignment issue to some extent.
Finally, 
experimental results on three palmprint datasets and ablation analysis show the superiority of our method.

To summarize, the contribution of our paper is four-fold:
\begin{enumerate}
	\item We have established a large-scale touchless palmprint dataset containing 2334 palms from 1167 individuals, 
	which will benefit the development of the community.
	\item We propose a novel deep learning framework comprising novel curved Gabor filters and well-designed 3D convolution, 
	which foucus on line features and can generate a high discriminative palmprint descriptor. 
	\item We develop a training supervision scheme that considers both the global descriptor and local descriptors,
	employing a block loss to enforce their robustness and discriminative ability.  
	\item Extensive experiments show that the designed CNNs achieve state-of-the-art performance, 
	proving the efficiency of palmprint recognition on large-scale person verification scenarios.
\end{enumerate}
  
The organization of this paper is shown as follows. 
The related works about Gabor filter, 
touchless palmprint dataset and recognition methods are briefly described in Section~\ref{sec_rw}.
The newly established dataset are introduced in Section~\ref{sec_dataset}.
Some visualization results are shown in Section~\ref{sec_motivation} and the motivation behind our method is also introduced.
Our proposed framework 3DCPN as well as its components and inference are analyzed in Section~\ref{sec_method}, 
followed by the experimental analysis in Section~\ref{sec_experiment}. 
This paper is finally concluded in Section~\ref{sec_conclusion}.

\section{Related Works}
\label{sec_rw}
\subsection{Gabor Filter}
Gabor wavelet is first invented by Dennis Gabor~\cite{gabor1946}, 
which adopts complex functions to describe the local information.
It can extract spatial local frequency features serving as an effective tool for texture detection.
For its significant visual properties, 
such as steerability and boundedness, 
the derived 2D Gabor filter is widely applied in palmprint recognition~\cite{Zhang2003Online,Genovese2019palmnet,luan2018gabor,zhao2020deep},
which is formulated as:
\begin{equation}
\begin{aligned}
	G(x,y;\lambda,\sigma,\theta,\gamma) &= \exp \left( -\frac{{x^\prime}^2+\gamma^2 {y^\prime}^2}{2\sigma^2} \right)
	\cos \left(2\pi \frac{x^\prime}{\lambda} \right) \\
	\label{eq_gabor}
\end{aligned}
\end{equation}
where $\lambda$ is the wave length, $\theta$ is the direction angle, 
$\sigma$ is the standard deviation of the Gaussian envelop and $\gamma$ determinates the aspect ratio. 
In Eq. \eqref{eq_gabor}, $x^\prime, y^\prime$ are projected coordinates with angle $\theta$.
\begin{equation}
\begin{aligned}
	x^\prime &= x\cos(\theta) + y\sin(\theta) \\
	y^\prime &= -x\sin(\theta) + y\cos(\theta)\\
\end{aligned}
\end{equation}

\begin{figure}[t]
	\centering
	\includegraphics[width=\linewidth]{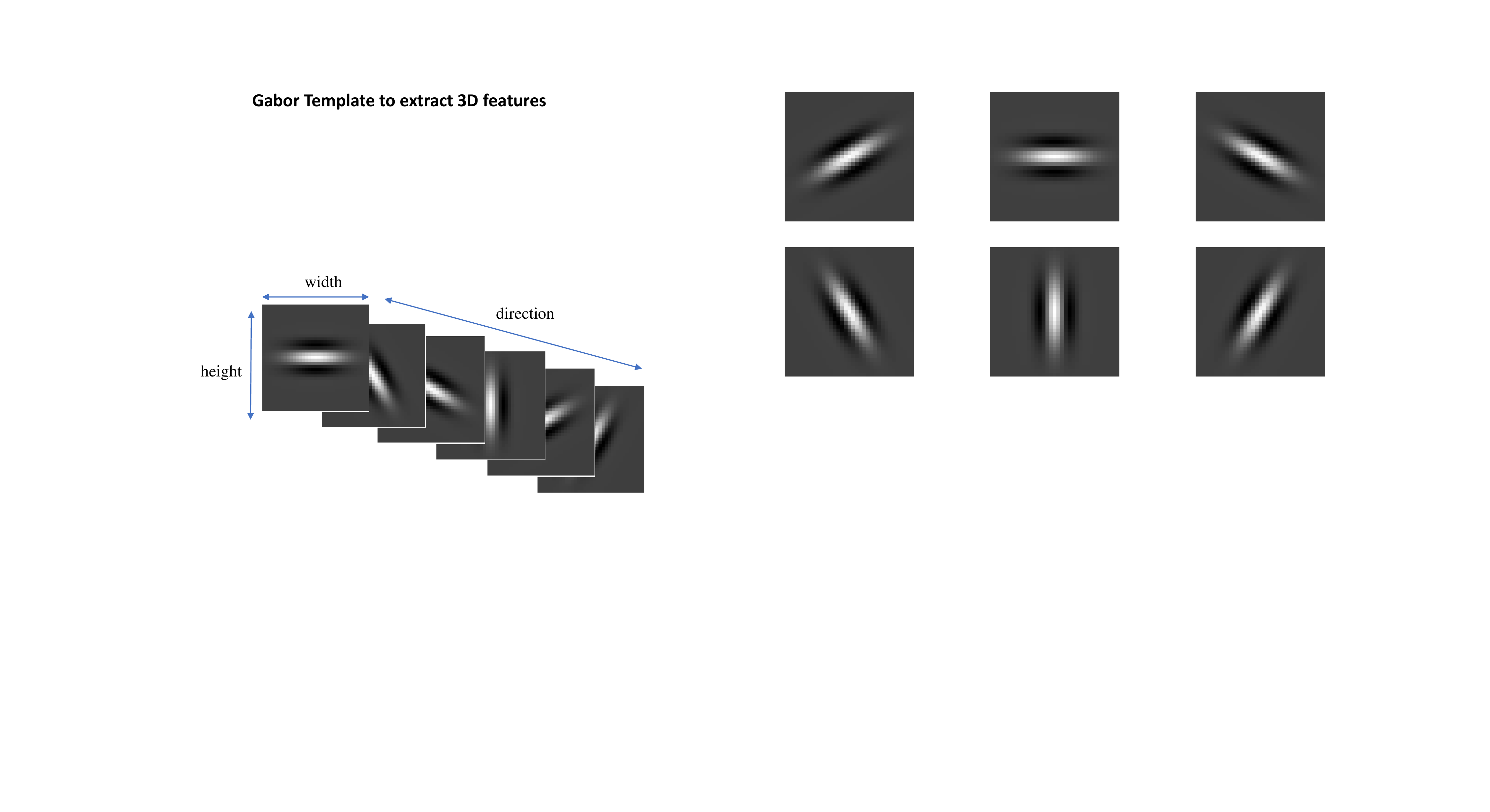}
	\caption{Gabor Template for extracting 3D features.}
	\label{fig_gabor} 
\end{figure}

In this paper, 
$\gamma$ is fixed to be 0.5 for enhancing line features and we regard the direction $\theta$ as one spatial dimension.
Hence, 
the Gabor filter could be denoted as:
\begin{equation}
\begin{aligned}
	G_t(x,y,\theta;\lambda,\sigma) = G(x,y;\lambda,\sigma,\theta,2.5)
	\label{eq_gabor_temp}
\end{aligned}
\end{equation}

where $\theta$ now represents the third dimension (depth dimension). 
The 3D filter $G_t$ is termed as \emph{Gabor template}.
A set of Gabor template $G(\lambda,\sigma)$ are shown in Fig. \ref{fig_gabor}.
In our framework,
we use multi-scale and multi-direction Gabor filters,
which are embedded into the first layer,
to extract low-level curve features.
In this way, 
the steerable properties are inherited into CNN 
and hence the model is robust to scale and orientation variations in palmprint images.

\begin{table*}[t]
\caption{Touchless palmprint datasets}
\begin{center}
	\scalebox{1}[1]{
	\begin{tabular}{|c|c|c|c|c|c|c|}
	\hline
	Dataset									        & Year	& Hands	& Images	& Image Size 			& ROI Provided 	& ROI Size \\
	\hline
	\hline
	CASIA\cite{CASIA}						    	& 2005	& 624 	& 5,502 	& 640$\times$460		& No			& NA	\\
	\hline
	IITD-v1\cite{IITD}						  		& 2006	& 460 	& 3,290 	& 1200$\times$640		& Yes			& 150$\times$150	\\
	\hline
	COEP\cite{COEP}							    	& 2010	& 168 	& 1,344 	& 1600$\times$1200		& Yes			& 512$\times$512	\\
	\hline
	TongJi\cite{zhang2017towards}					& 2017	& 600 	& 12,000 	& 800$\times$600		& Yes			& 128$\times$128	\\
	\hline
	NTU-CP-v1\cite{matkowski2019palmprint}			& 2019	& 655 	& 2,478 	& Mdn. 1373$\times$1373	& No			& NA	\\
	\hline
	XJTU-UP\cite{shao2020towards}					& 2020	& 200 	& $>$20,000 & From 3264$\times$2448 to 5312$\times$2988		& Yes			& 128$\times$128	\\
	\hline
	CUHKSZ-v1								        & 2021	& 2334 	& 28,008 	& 1024$\times$768		& Yes			& 128$\times$128	\\
	\hline
	\end{tabular}}
\end{center}
\label{tab_dataset}
\end{table*}

\subsection{Palmprint Recognition Methods}
Based on how the feature extraction kernels are obtained,
the approaches in palmprint recognition could be roughly categorized into conventional methods and deep learning based methods.
In this section, 
we first review several popular coding algorithms and then introduce recent advancements in deep learning methods. 

\subsubsection{Conventional Methods}
Starting from PalmCode~\cite{Zhang2003Online} and Competitive Code~\cite{Kong2004Competitive},
coding methods~\cite{jia2008palmprint} have shown its superiority both in speed and accuracy. 
In these approaches, 
line features are extracted by human-designed filters and the orientation information is then encoded into the feature vector.
Incorporating fast implementation in programming, 
these methods achieve high performances in person verification.
In DOC~\cite{fei2016double},
two orientation information is utilized
and an improving nonlinear matching algorithm is designed.  
Instead of comparing code features by pixels,
Luo \emph{et al.}~\cite{luo2016local} proposes a matching algorithm at region-level named LLDP.
The feature map is first split into several grids and a histogram-based distance is calculated in the matching. 
An improved local binary descriptor is proposed in~\cite{fei2019localdiscriminant} that combines the information extracted from an exponential Gaussian fusion model.
In~\cite{fei2019localapparent},
both latent direction code and apparent direction code are extracted and further leveraged in the histogram matching.
CR\_CompCode~\cite{zhang2017towards} is a popular learning based method that leverages the training gallery information.
The proposed collaborative representation obtains high recognition accuracy while having an extremely low computational complexity. 
Fei \emph{et al.}~\cite{fei2016low} leverage low-rank representation to conduct subspace clustering of palmprint images,
which is also robust to noisy images.
In~\cite{jia2017palmprint},   
a general framework for direction representation based method is proposed,
consisting of strategies of multi-scale,  
multi-direction level, multi-region, and feature selection. 
Multiple features are fused according to the correlation and redundancy among them. 
Recently, 
Fei \emph{et al.}~\cite{fei2019learning} propose a binary code learning model that can generate discriminant direction feature maps for accurate palmprint matching. 
SIFT~\cite{Lowe04distinctiveimage} is an effective local descriptor that can detect keypoints and is scale-invariant.
For alleviating the common alignment issue in palmprint image matching, 
Zhao \emph{et al.}~\cite{zhao2013contactless} adopts SIFT features and a novel iterative RANSAC (Random Sample Consensus) algorithm.
With the aid of the scale invariance property of SIFT descriptor, 
the approach improves the verification accuracy on many datasets.
Charfi \emph{et al.}~\cite{charfi2016local} combines SIFT and sparse representation method, 
perform fusion on the left and right palm features at rank level using Support Vector Machines (SVM) classifier.

\subsubsection{Deep Learning Methods}
With the significant achievement of deep learning in computer vision, 
many methods in palmprint recognition~\cite{Dian2016Contactless,zhao2019joint,zhao2020deep,Jia2021Performance} apply convolution neural networks (CNN) as a key component for feature extraction. 
Zhao \emph{et al.}~\cite{zhao2019joint} uses stacked CNNs for hyperspectral palmprint feature extraction where input images are formatted as a cube.
Zhang \emph{et al.}~\cite{Zhang2018Palmprint} proposes an architecture based on Inception network, 
which could be applied for both palmprint and palm vein feature extraction.
In order to realize efficient palmprint feature matching,
Shao \emph{et al.}~\cite{Shao_2019_CVPR_Workshops} combines hash coding and knowledge distillation via a compressed deep neural network.
In~\cite{matkowski2019palmprint}, 
an end-to-end deep learning algorithm is proposed for accurate palmprint identification.
The whole network consists of two parts, 
one pretrained VGG network is designed for palm alignment and detection, 
and another part is responsible for feature extraction. 
Genovese \emph{et al.}~\cite{Genovese2019palmnet} adopts adaptive Gabor filters and Principal Component
Analysis (PCA) in a three-layer CNN that can learn high-level features from heterogeneous devices.
In ~\cite{Zhao2020Joint}, 
a joint constrained least-square regression (JCLSR) framework is designed, 
which learns local region representation by CNNs that requires all local regions have similar projected target matrice.
Based on fully convolutional network design, 
Liu \emph{et al.}~\cite{liu2020contactless} employs residual blocks to extract high discriminative features.
A variant of triplet loss~\cite{hoffer2015deep} is adopted to enhance distance distribution.
Zhu \emph{et al.}~\cite{zhu2020boosting} also uses a metric learning method which makes the feature distribution more uniform.

\subsection{Touchless Palmprint Datasets}
\label{sec_rw_dataset} 
With the development of touchless palmprint recognition,
many researchers have established touchless palmprint image datasets.
In this section, 
we put emphasis on the popular 2D image datasets that are widely used in the community.
The earliest touchless palmprint dataset is released by the Chinese Academy of Sciences (CASIA)~\cite{CASIA}
and by the Indian Institute of Technology in Dehli (IITD-v1)~\cite{IITD}.
They both use digital cameras to capture hand images in a stable environment. 
CASIA contains 5,502 palmprint images captured from 624 hands and IITD contains 3,290 images from 460 hands.
In 2010, the College of Engineering Pune (COEP)~\cite{COEP} released a high-quality
touchless palmprint dataset which is acquired in a more constrained environment.
The hands are put in a semi-box and there are pegs to guide the positions of fingers.
\cite{zhang2017towards} also uses a semi-box design and collects 1,2000 images from 600 palms.
Recently, 
\cite{matkowski2019palmprint} release a dataset NTU-CP-v1 of palmprint images where the hand pose varies considerably. 
The dataset contains 655 classes with 2,478 hand images in total.
The details of the touchless datasets are summarized in Table \ref{tab_dataset}. 

\section{CUHKSZ-v1 Dataset}
\label{sec_dataset}
\subsection{Acquisition Device}
\begin{figure}[t]
\centering
\includegraphics[width=\linewidth]{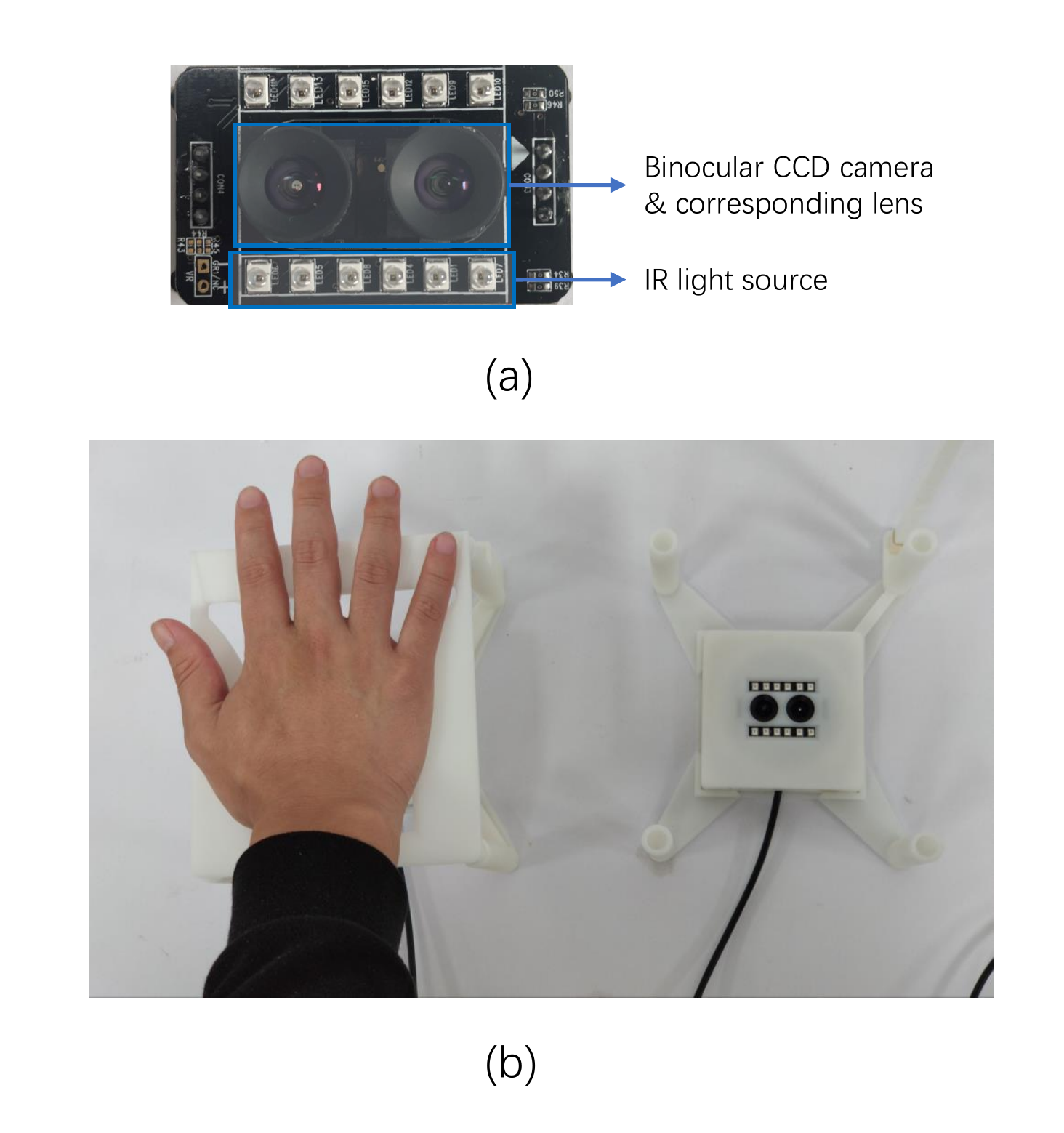}
\caption{Palmprint acquisition devices. 
	(a) is the structure of the acquisition module,
	which consists of one binocular camera and an IR light source. 
	The binocular camera can capture palmprint and palm vein images simultaneously.
	(b) shows the acquisition process. 
	First, the volunteer puts their hand on the left device (Device1) with a fixed height.
	Then, he/she needs to put their hand on the right device (Device2) with different heights.}
\label{fig_device} 
\end{figure}

While the palmprint image capturing process is simple, 
the acquisition device should be flexible, user-friendly, and can acquire high-quality images. 
In order to collect large-scale touchless palmprint dataset in an open environment,
we design a simple palmprint acquisition module.
As shown in Fig. \ref{fig_device},
the module contains a binocular CCD camera (camera1 and camera2), camera lens, a USB cable, 
and an IR light source.
With the IR light, 
the binocular camera can capture palmprint and palm vein images at the same time.
The resolution of captured images is 1024$\times$768.
The whole module is encased in a plastic shell and placed in a bracket,
composing the whole acquisition device.
Since the field of the view and the focal length of the camera are fixed,
we set a height reference to help volunteers put their hands on an appropriate height.
For the same reason, an optional panel can be added on the top of the bracket.
The simple structure of our newly designed device improves ease of use. 
Different from semi-closed box designs~\cite{kumar2008incorporating,zhang2017towards}, 
our acquisition system only has a height reference which is highly user-friendly.
 
\begin{figure}[t]
\centering
\includegraphics[width=\linewidth]{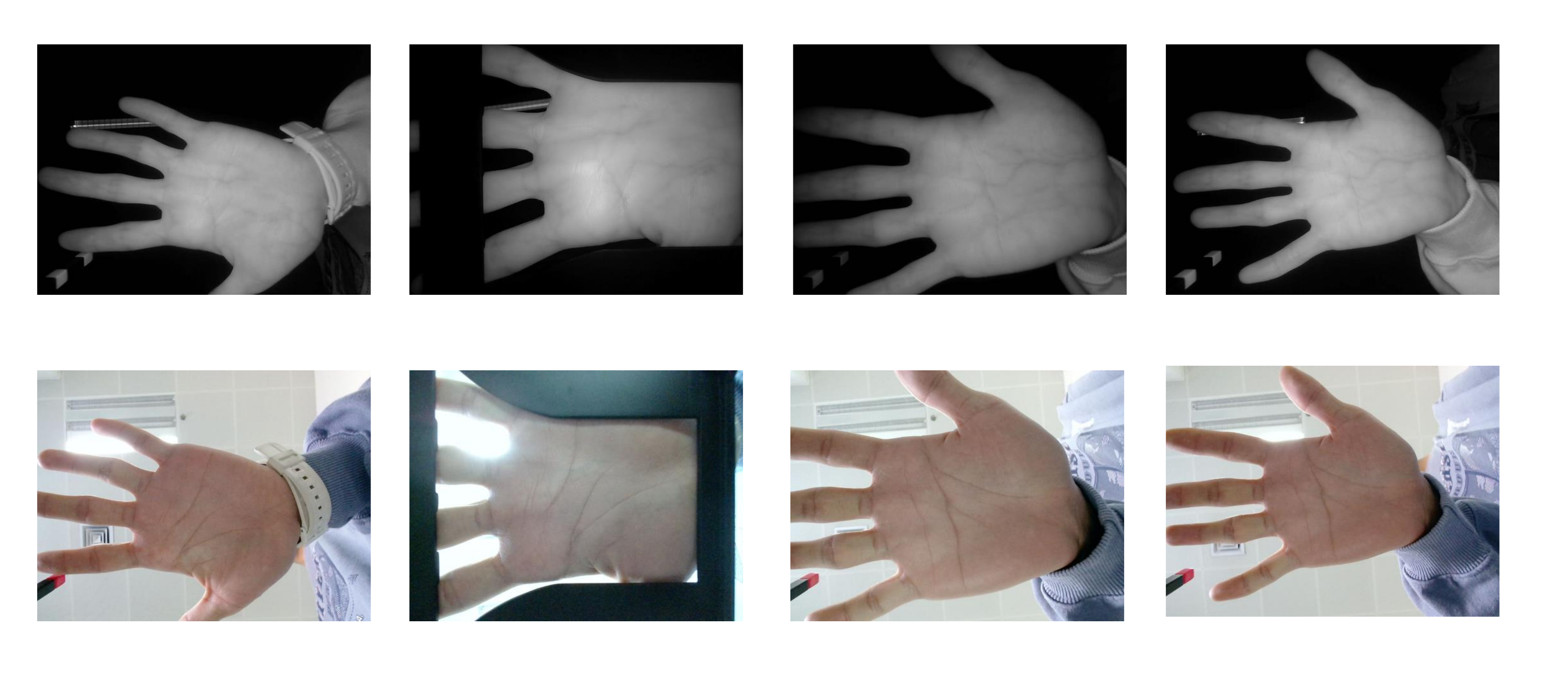}
\caption{Examples of the captured images of one individual. 
		The first row shows the palm vein IR images and the second row is the corresponding palmprint RGB images.}
\label{fig_images} 
\end{figure}

\subsection{Data Acquisition Process} 
\label{data_acquisition}
Generally, 
the palmprint verification process in real life has two stages. 
People first register their palmprint images on the device and they can authenticate identity after by showing their hands. 
To simulate the real application,  
we also set two stages for acquiring palmprint images corresponding to the acquisition devices Device1 and Device2,
as shown in Fig. \ref{fig_device}(b). 
The only difference between the devices is that Device1 has the panel on the top, 
which can ensure the hand is placed at a fixed height.
In stage one, 
people are asked to place their hand on Device1 and each palm is captured 3 times for registration.
So for one individual,
there are 6 RGB images and 6 IR images from 2 palms.
In stage two,
people repeat the above steps but the height of the hand varies in each capture to simulate real application.
The images captures in this stage are used for test.
Then after the acquisition,
we have 12 RGB images and 12 IR images from 2 palms for each individual.
In our dataset, images were collected from 1167 volunteers from the university, 
comprising total 14004 palm images.  
In Fig. \ref{fig_images} we show examples of captures images of one person.
During the acquisition process,
we did not place much constraints on the hand pose.
As our device is straightforward to use,  
we only reminder the volunteer to stretch his/her fingers during acquisition.

\begin{figure}[t]
	\centering
	\includegraphics[width=\linewidth]{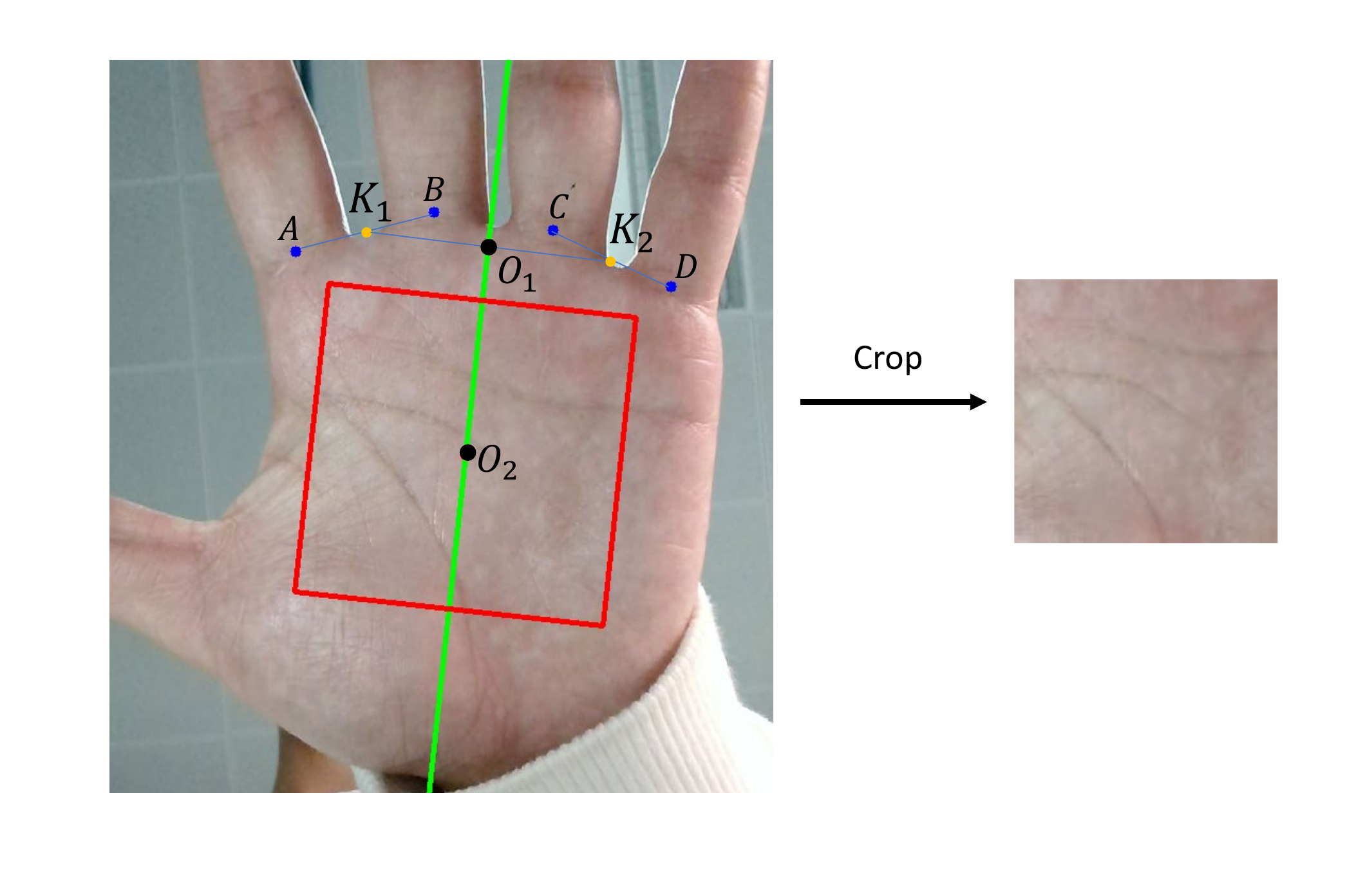}
	\caption{The illustration of label annotation and ROI extraction. 
	We first label four keypoints $A,B,C,D$ on the joints between fingers and palm to determine the finger gaps $K_1, K_2$.
	Then we build a local coordinate system based on the finger gaps.
	The ROI region (red box) is finally located based on the local coordinate system.
	}
\label{fig_annotate} 
\end{figure}

\begin{figure*}[t]
	\centering
	\includegraphics[width=\textwidth]{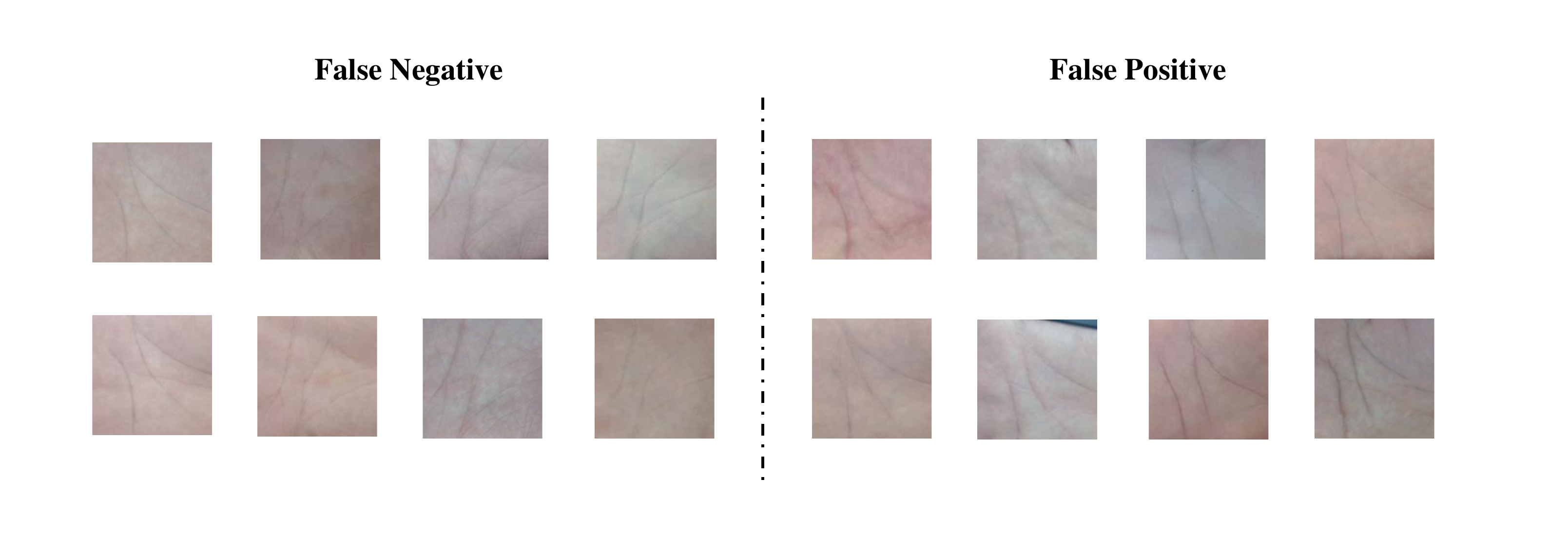}
	\caption{Hard examples. 
	A main issue of false negative pairs is principal line misalignment and 
	the false positive pairs all have similar line features that cause the mismatch.
	}
\label{fig_hard} 
\end{figure*}
 
\begin{figure}[t]
	\centering 
	\includegraphics[width=\linewidth]{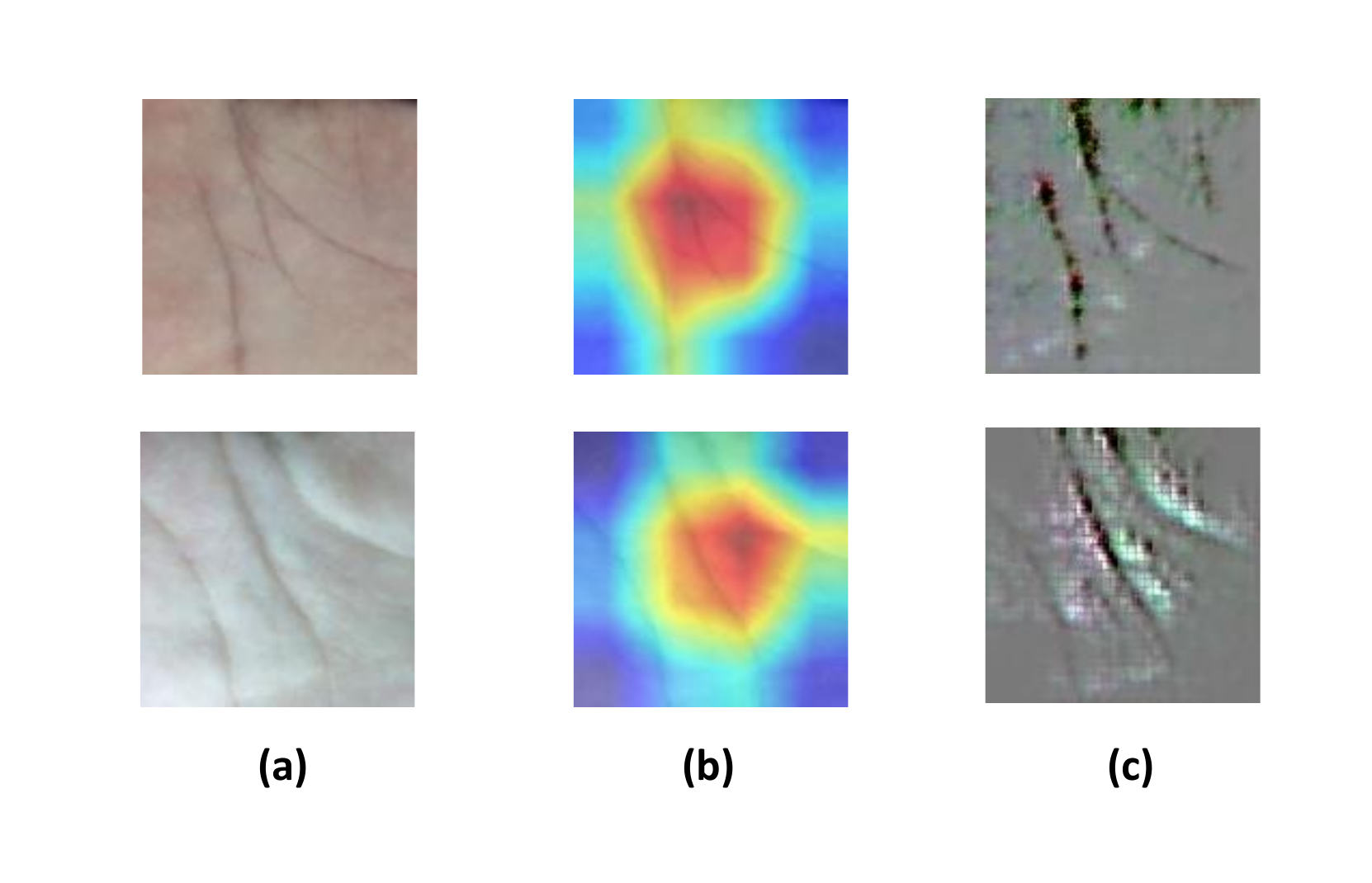}
	\caption{Visualization of learning heat maps. 
	Column (a) is the original ROI, 
	column (b) is the class activation map and column (c) depicts the guided backpropagation map.
	}
\label{fig_grad} 
\end{figure}  

\subsection{Keypoints Annotations and ROI Extraction}
Most existing works follow~\cite{Zhang2003Online} to determine the region-of-interest (ROI) region, 
which is based on finding two finger gaps (the gap between the index and the middle fingers and the gap between the ring and the little fingers).
We also follow this scheme but take a more robust way.
Instead of annotating the finger gaps directly, 
for each image we label the four joints between fingers and palm.
Then the finger gaps are naturally the two midpoints of the connecting lines. 
The scheme can decrease the bias since the finger gaps are not easy to precisely located when the hand postures vary.
In addition, 
the ROI is determined by four points which further improve the stability of ROI localization.
The influence of the ROI bias is analyzed in Section \ref{sec_bias}.
Though we didn't annotate the palm vein images,
their ROIs could be further annotated in the same way or be extracted by alignment methods.

Concretly, as shown in Fig. \ref{fig_annotate}, 
from left to right we denote the four keypoints as $A,B,C,D$.
The finger gaps $K_1,K_2$ are the midpoints of $AB$ and $CD$ respectively.
$O_1$ is the midpoint of $K_1K_2$.
We build a local coordinate system with the origin $O_1$ and the X-axis $O_1K_1$.
Denote the distance between the finger gaps as $l=||K_1K_2||$. 
$O_2$ is the center of the palm which lies on the Y-axis and $||O_1O_2|| = 0.85l$.
Centered on $O_2$,
the ROI region (red box) is a squared box parallel to the coordinate system.
The length of the ROI box is $1.25l$.

\begin{figure*}[t]
	\centering
	\includegraphics[width=\textwidth]{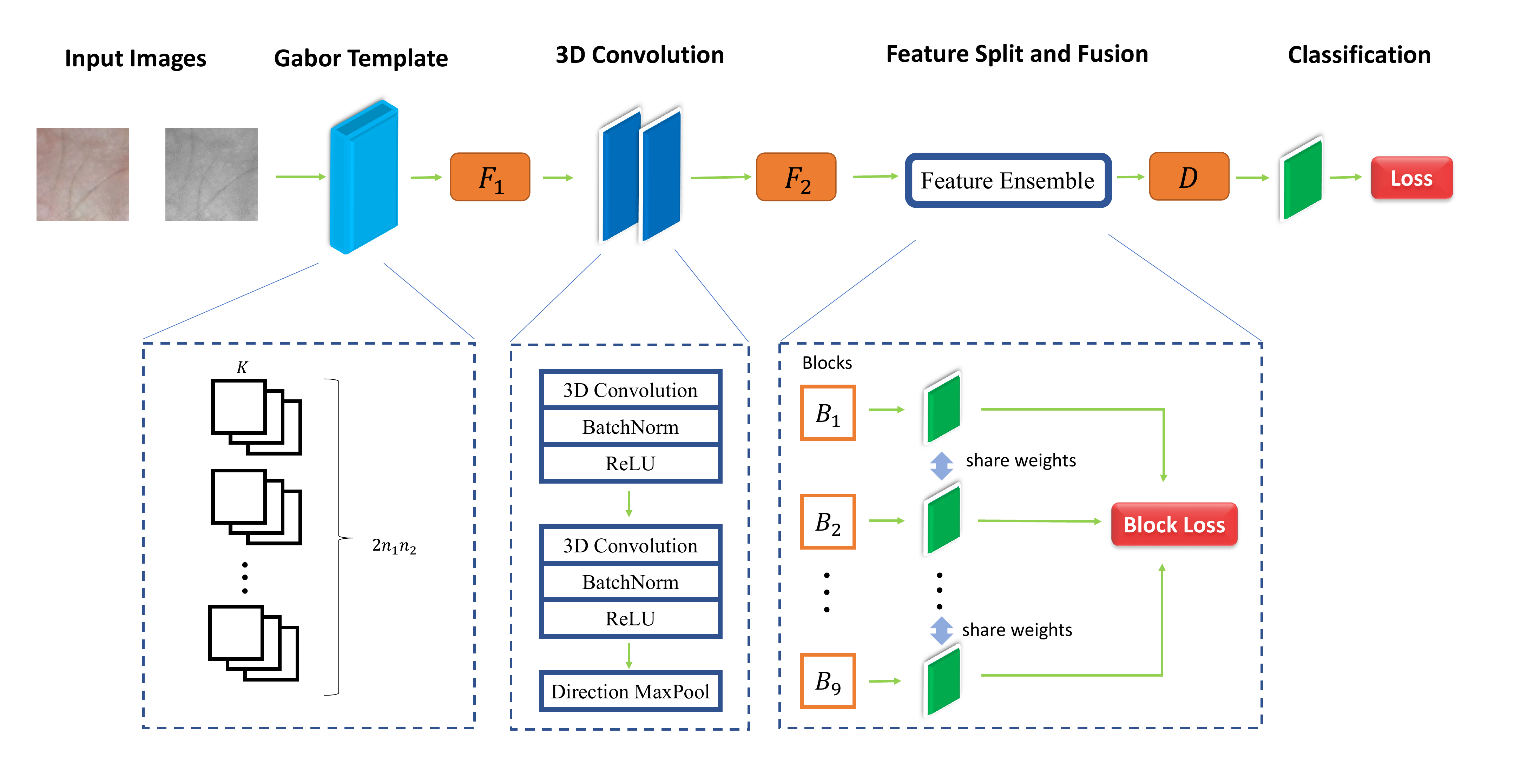}
	\caption{Overview of our proposed 3DCPN framework. 
	The palmprint is first converted to a gray-scale image and input to the neural network.
	In the first layer of the 3DCPN, 
	the Gabor template is employed to extract 3D features $F_1$ which are fed into two 3D convolution modules to extract high-level feature $F_2$.
	Then the block features are split from $F_2$ which are supervised by the proposed block loss.
	The final palmprint descriptor $D$ is ensembled from the block features by dynamic fusion.
	On the top of the whole network, a classification loss is added (Best view in color).}
\label{fig_pipeline} 
\end{figure*}

\section{Motivation}
\label{sec_motivation}

In this section, 
we provide some visual results of palmprint feature learning with neural networks,
and introduce the intuition behind our line-based alignment method.
Specifically, 
we conduct palmprint ROI classification experiments with several baseline networks including ResNet18, GoogLeNet and VGG11-bn. 
In the training, 
the ROI images are input directly to the networks which are supervised by single softmax loss.
We carefully tune the parameters in the experiments so that the results could well reflect the feature learning process.
We observe similar phenomenons with different network architectures and exhibit the results of ResNet18 below as a representative example.

\subsection{Role of Principal Lines} 
For face recognition task,
deep learning methods usually relies on extracting high-level features (nose, eyes, \textit{i.e.}).
However, 
the discriminative information on palmprint lies in line features which is low-level~\cite{zhang2017towards}.
To locate the learned part in palmprint ROI, 
we adopt GramCAM~\cite{Selvaraju2017Grad} and Guided Backpropagation~\cite{SpringenbergDBR14} to show attention maps as shown in Fig. \ref{fig_grad}.
In the figure, 
column (b) is the class activation map representing where the model has to focus on in the inference.
Column (c) depicts the guided backpropagation map corresponding to positive gradient with the correct class.
We can see from the figures that the principal lines are the most distinctive parts in palmprint ROI,
which is consistent with the claims in traditional methods~\cite{Kong2004Competitive,zhang2017towards}.
In addition, 
another observation is that the wrinkle lines on palm are generally neglected in the learning process because the network tends to learn simple features first.
In addition, 
the wrinkle lines are not always clear in palmprint capture with touchless manner.
Similar to traditional coding based methods,
this principal line based learning scheme makes CNN sensitive to spatial variance,
which is also demonstrated in hard examples. 

\subsection{Hard Examples} 
Knowing that the key part is principal lines, 
we enumerate hardest mismatched ROI paris in Fig.~\ref{fig_hard} to analyze the performance gap.
When we compare the principal lines of each pair, 
it is apparent that a main issue of false negative pairs is principal line misalignment due to hand distortion or pose variance.
Likewise, 
the false positive pairs all have similar line features that cause the mismatch.
The results show that the ROI misalignment can degrade the recognition performance of CNN largely.
In real application,
the non-unified localization scheme of palmprint ROI across datasets could make the problem more serious.

\section{3DCPN}
\label{sec_method}

In this paper, 
we focus on the feature extraction of palmprint ROI images and use directly the ground truth ROIs,
while the palm detection task is left for further research.
The goal of our method is to extract high discriminative features for palmprint recognition,
especially on large-scale datasets.
According to the analysis before, 
we want the neural network to focus on line features,
and the feature should be robust to possible ROI localization bias to achieve high verification performance.
To fulfill the task, 
we propose a novel framework named 3DCPN which leverages low-level Gabor features.
The overall pipeline is shown in Fig. \ref{fig_pipeline}.
In the first layer,
we combine conventional Gabor filter and curved Gabor filter to extract Gabor features $F_1$.
Then two 3D convolution modules are employed to ensemble the extracted low-level features,
which yield feature map $F_2$.
After pooling $F_2$ on the depth dimension (direction dimension), 
the orientation information is further enhanced.
Finally we obtain the palmprint descriptor $D$ by spatial split and dynamic concatenation.
The whole network is supervised by a classification loss and our designed block loss.

\begin{figure}[t]
	\centering
	\includegraphics[width=\linewidth]{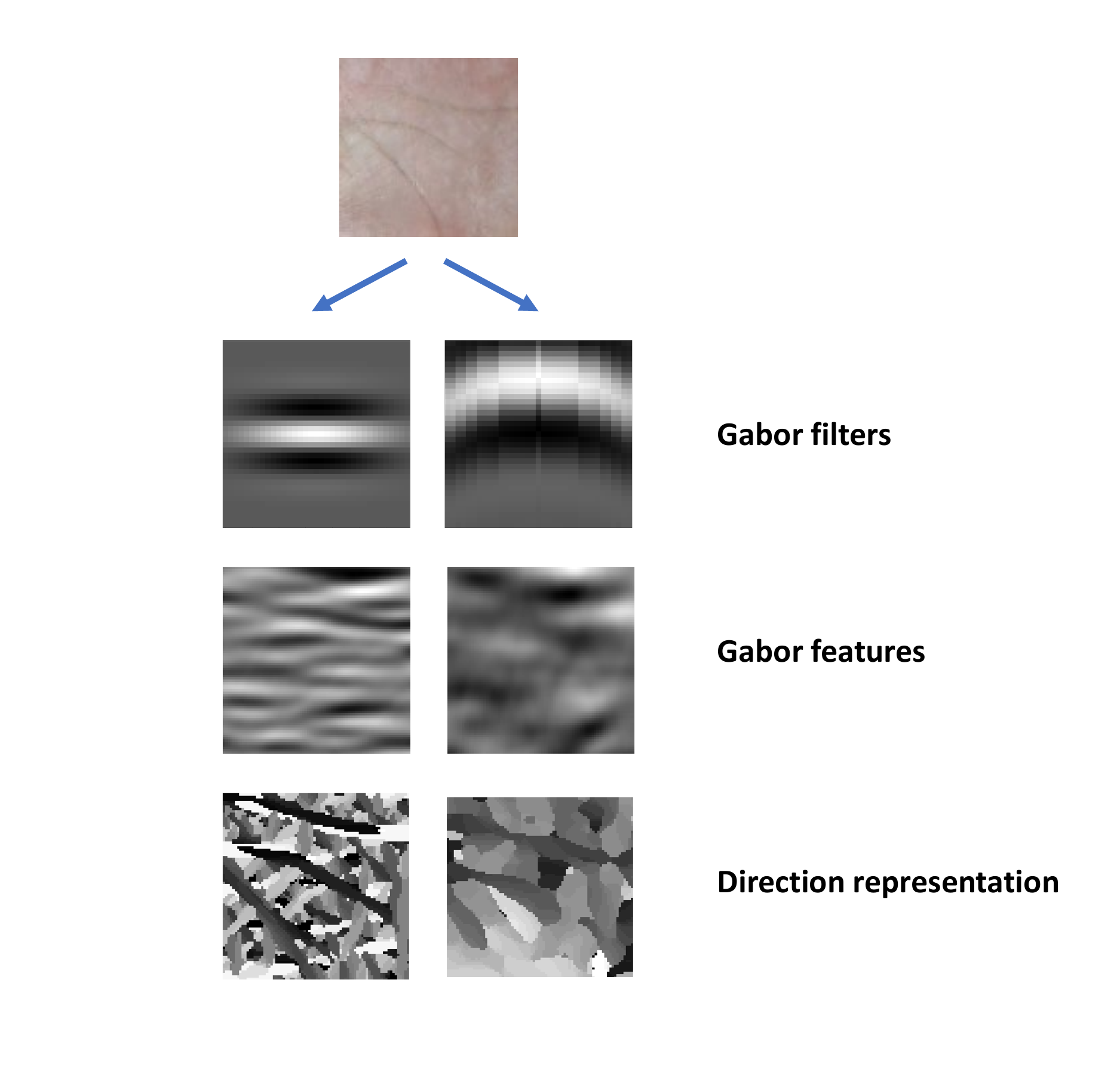}
	\caption{The illustration of difference between Gabor filter and curved Gabor filter.
	The proposed curved Gabor filters extract curve features in the palmprint and produce extra information.
	}
\label{fig_cgabor} 
\end{figure}

\subsection{Curved Gabor Template} 
The Gabor filter is a powerful line feature extractor,
however it is not designed for detecting non-straight curves in images.
A palmprint image may have various curves and Gabor filters are not sufficient to describe these features.
To bridge the gap,
we generalize the original Gabor template $G(x,y,\theta)$, 
which is named as curved Gabor template, 
by mapping the pixel values in the wavelet line to a predefined curved curve.
The mapping shifts the pixels in $G(x,y,0)$ and the shift distance forms a (part) circle, 
which we called shifting circle, as shown in Fig. \ref{fig_cgabor}.
Specifically, 
the shifting circle is centered on the midpoint of the filter border and its radius is one-third of the filter size.
Then we rotate the shifted filter by $\theta$ to obtain the curved Gabor template $G_c(x,y,\theta)$.

Obviously,  
each $\theta$ in the curved Gabor template corresponds to one 2D filter which we term as a curved Gabor filter.
In fact, the curved Gabor filter extends line feature types extracted from palmprint images.
From Fig. \ref{fig_cgabor} we can see that circular curves or similar curvature have a high response when convolves with a curved Gabor filter.
We will show the complementarity of two kinds of Gabor filters in Section \ref{sec_exp_gabor}.

\subsection{Low-level Feature Extraction with 3D Convolution Layer}
In Eq. \eqref{eq_gabor_temp},
the shape of the wavelet is controlled by $\sigma$ and $\lambda$ ($\gamma$ is fixed).
A larger $\lambda$ generates wider wavelets and $\sigma$ is a scale-related parameter that determines the wavelet length.
A combination of multiples Gabor templates can detect a wide range of line features,
providing abundant information for the network.

In 3DCPN,
multi-scale and multi-direction Gabor filters are embedded in the first layer to extract Gabor features,
which are generated by following rules.
Given the candidate sets of above hyperparameters $\Lambda = \{\lambda_1, \lambda_2, ...,\lambda_{n_1}\}$
and $\Sigma = \{\sigma_1, \sigma_2, ...,\sigma_{n_2}\}$,
the hyperparameters of the Gabor templates $G(x,y,\theta)$ and curved Gabor templates $G_c(x,y,\theta)$ are Cartesian product of the two candidate sets.
Let the number of orientations be $K$,
we then have totally $2 \cdot n_1 \cdot n_2 \cdot K$ Gabor filters in the first layer.
After passing the first layer,
the Gabor features $F_1 \in \mathbb{R}^{2n_1 n_2 K \times H_1 \times W_1}$ are extracted,
where $2n_1 n_2 K$ is the number of output channels and $H_1,W_1$ are output height and width.
It should be pointed out that we regard the orientation as depth dimension,
therefore we rearrange $F_1$ to become a 3D feature map with the shape $2n_1n_2 \times H_1 \times W_1 \times K$.
All the parameters in the first layer are handcrafted and froze that will not update in the training process.

After the first layer, 
we place two 3D convolution modules to ensemble $F_1$ in three spatial dimensions.
Each module contains a 3D convolution layer, 
a BatchNorm layer~\cite{10.5555/3045118.3045167} and ReLU activation. 
The convolution strides are (2,3,3) and (1,1,1) respectively.
The output channels of the 3D convolution layers are $8 \cdot n_1 \cdot n_2 \cdot K$ which is 4 times larger than the first layer.
In accordance with previous works~\cite{Zhang2003Online,zhang2017towards} that apply local dominant orientation code,
the depth dimension is max pooled to 1.
Convolution layers conserve local information in the feature map to the most extent,
making it possible to adopt local losses described in the next section.
For the same reason, 
we abandon the pooling layer in height and width dimensions.
 
\subsection{Spatial Split and Block Loss}
Local information is critical for coding based methods~\cite{luo2016local,fei2019learning} 
that usually rely on per-pixel matching or local histogram matching.
Though the distortion and misalignment often occur on palmprint images 
(due to camera configuration, relative hand position, hand posture \textit{etc.}),
strong spatial relations still exist among different regions on images.
A basic principle is that the same part on two images of one palm should also match.
Moreover, 
the region-based recognition scheme could also alleviate the ROI misalignment issue in the matching. 

Under the above intuition, 
we split the feature map into several local blocks and design our local block loss.
To be specific,  
we first pad $F_2$ such that its width $W_2$ and height $H_2$ are divisible by 3 and then split the feature map $F_2$ into $3 \times 3$ parts with equal size.
Denote the split blocks as $B_i \in \mathbb{R}^{\frac{H_2}{3}\times \frac{W_2}{3}}, i=1,2,...,9$.
$B_i$ contains local information and should be discriminative, 
so we optimize the classification loss on each of these spatial block features separately.
As a result,     
block feature $B_i$ is trained to be distinctive.
After forwarding each $B_i$ to a shared classification layer (fully-connected layer), 
the block loss $L_{B}$ is calculated as:

\begin{equation}
\begin{aligned}
	L_B &=  \sum_{i}^{9}L_{cls}(FC_1(B_i), y)\\
\end{aligned} 
\end{equation}
where $L_{cls}$ represents softmax loss or its angular margin version~\cite{deng2019arcface},
$y$ is the ground truth class label.

The final palmprint descriptor $D$ is ensembled by block features.
On one hand,
it should contain global information which is used to distinguish identity.
On the other hand, 
it learns and controls the importance factor of each block.
Here for obtaining the palmprint descriptor $D$, 
we adopt a weighted summation strategy to fuse the block features.
Considering that different regions may contribute unequally to the final fused descriptor,
a learnable weight parameter $W = \{ w_1, w_2, ..., w_9\}$
is set to fuse the block features dynamically:
\begin{equation}
\begin{aligned}
	D = FC(\sum_i^{9} \frac{\exp(w_i)}{\sum_k^{9}\exp(w_k)}B_i).
\end{aligned}
\end{equation}
where $FC$ is a fully-connected layer and $W$ is normalized by softmax operation such that the overall coefficient sums to 1.
After, 
the descriptor is flattened to a single vector $D \in \mathbb{R}^{1024}$.

\subsection{Network Supervision and Feature Matching}
The principle loss function is a classification loss $L_{D}$ added on the top of the network to supervise descriptor $D$:

\begin{equation}
\begin{aligned}
	L_D &=  L_{cls} \left(FC_2(D), y \right)
\end{aligned}
\end{equation}
where the classification layer has output channels that equal to the number of classes. 
With the predefined block loss, 
the total loss of the framework is formulated as:
\begin{equation}
\begin{aligned}
	L_{total} &= L_D + \mu L_B \\
\end{aligned}
\end{equation}
where $\mu$ is the trade-off loss weight.
As to the choice of classification,
we test two popular loss functions, 
softmax loss and arc-margin loss~\cite{deng2019arcface}. 
The formulation of the two loss functions are:
\begin{equation}
	\begin{aligned}
		L_{softmax} &= -\frac{1}{N} \sum_{i=1}^N \log \frac{e^{z_i}}{\sum_{j=1}^n e^{z_j}} \\
	\end{aligned}
\end{equation}
\begin{equation}
	\begin{aligned}
		L_{arc} =& -\frac{1}{N} \sum_{i=1}^N \log \frac{e^{s\cos(\theta_{y_i}+m)}}{e^{s\cos(\theta_{y_i}+m)}+\sum_{j=1,j\neq y_i}^ne^{s\cos(\theta_{j}+m)}} \\
	\end{aligned}
	\label{eq_arc}
\end{equation}
where $z_i$ is the output logit of the $i$-$th$ sample, 
$y_i$ denotes its label,
$n$ is the number of classes and $N$ is the batch size.
In Eq. \eqref{eq_arc},
$\theta_i$ is the angle between $i$-$th$ logit and weight in classification layer.
$s$ and $m$ are hyperparameters which represent scale and angular margin respectively.

In this paper,
We choose the arc-margin loss (Eq. \eqref{eq_arc}) for palmprint verification since it can achieve higher performance and is widely applied in verification tasks.
The comparison of the two above loss functions is discussed in Section \ref{sec_arc}.
For palmprint matching,
the descriptor $D$ is used and the cosine distance is leveraged. 

\section{Experiments}
\label{sec_experiment}

\begin{table*}
  \caption{The performance comparison (\%) on touchless palmprint datasets}
	\centering
		\begin{tabular}{lccccccccccc}
		\toprule
		\multirow{2}{*}{Methods} && \multicolumn{2}{c}{CUHKSZ}  && \multicolumn{2}{c}{TongJi} && \multicolumn{2}{c}{IITD}\\
		\cmidrule(lr){3-4}  \cmidrule(lr){6-7} \cmidrule(lr){9-10}
						&& Rank-1 	& EER 	&& Rank-1   & EER	&& Rank-1	& EER \\
		\toprule
		CompCode 	  	&& 99.72	& 0.42 	&& 99.92 	& 0.10 	&& 98.11	& 2.79	\\
		OrdinalCode		&& 99.72 	& 0.38 	&& 99.89 	& 0.15 	&& 94.92 	& 4.17 	\\
		LLDP$_1$	  	&& 99.53 	& 0.56 	&& 99.92 	& 0.12	&& 98.03 	& 2.62	\\
		LLDP$_2$	  	&& 99.15 	& 0.78 	&& 99.85 	& 0.15	&& 97.46 	& 3.15	\\
		LLDP$_3$	  	&& 99.79 	& 0.32 	&& 99.89 	& 0.12	&& 96.61 	& 2.13	\\
		CR-CompCode 	&& 99.47	& 0.74 	&& 99.94 	& 0.06 	&& 94.42 	& 3.89	\\
		\hline
		Resnet18 	  	&& 98.24	& 0.87 	&& 99.85 	& 0.15 	&& 92.32	& 3.57	\\
		VGG11-bn 	  	&& 90.63 	& 2.40 	&& 99.14 	& 0.31	&& 93.21 	& 3.04 \\
		GoogLeNet 		&& 88.71 	& 2.77 	&& 98.67 	& 0.56	&& 94.73 	& 2.93 \\
		PalmNet 	  	&& 95.97 	& 0.76 	&& 99.83 	& 0.17 	&& 92.48 	& 4.46 \\
		BPFNet 	  		&& 99.83 	& 0.23 	&& 99.83 	& 0.24 	&& 96.87 	& 2.89 \\
		3DCPN		    && \textbf{100}	& \textbf{0.08}	&& \textbf{100} & \textbf{0.02} && \textbf{98.32}	& \textbf{2.55} \\
		\bottomrule
		\end{tabular}
\label{tab_test2register}
\end{table*} 

\begin{figure*}[t]
	\centering
	\includegraphics[width=\textwidth]{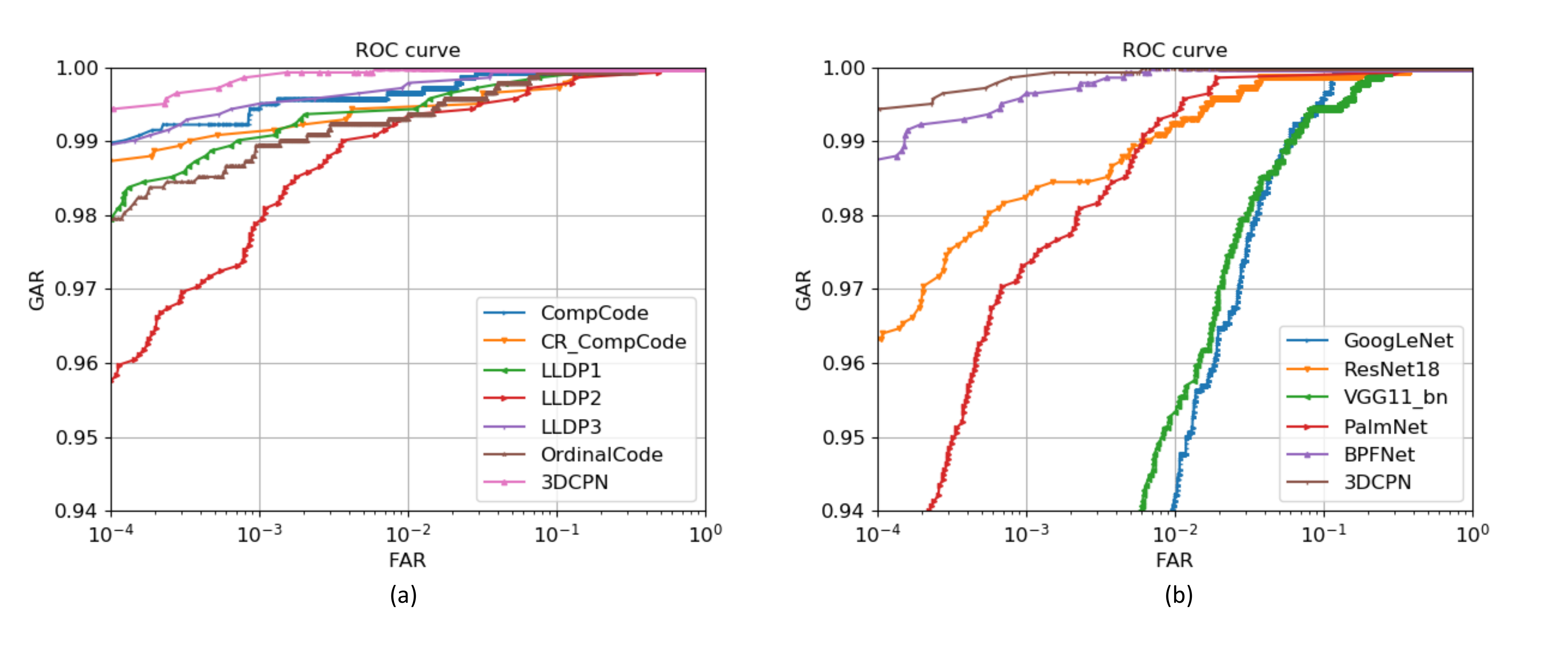}
	\caption{The ROC curves obtained using the methods listed in Table \ref{tab_test2register}.
			The figure shows the comparison of our 3DCPN with (a) coding based methods and
			(b) deep learning based methods.}
\label{fig_roc} 
\end{figure*}

In this section, 
we provide experimental results on several popular touchless palmprint benchmarks,
where the corresponding metrics are introduced.
We provide the comparison results with state-of-the-art methods as well as some baseline methods.
The robustness to possible ROI bias is also discussed.
Following we provide comprehensive experiments to discuss the effect of components in 3DCPN in ablation studies. 
We investigate the influence of parameter settings, 
including the arc-margin loss and the trade-off hyperparameter $\mu$, in the last part.

\subsection{Datasets and Metrics}
To evaluate our proposed method,
we conduct experiments on three touchless palmprint benchmarks,
including our newly collected dataset CUHKSZ-v1, TongJi, and IITD-v1. 
The details of the benchmarks are described in Section \ref{sec_rw_dataset}.

\subsubsection{Dataset Split} 
Concerning the dataset split,
in our CUHKSZ dataset 931 individuals are randomly chosen as training samples and the rest 236 individuals are test samples.
Each palm is regarded as one class, 
so there are totally 1862 classes in the training set and 472 classes in the test set.
For TongJi dataset, 
we follow the official train/test split and the details could be found in~\cite{matkowski2019palmprint,zhang2017towards}.
There is no official split for IITD~\cite{IITD} dataset,
in this paper the first 400 palms are split into training set and the rest 60 palms are in the test set.
No images from any database are discarded in the evaluation.
 
\subsubsection{Metrics}  
Because most existing palmprint recognition methods and touchless palmprint databases are designed for person verification rather than identification,
we use Rank-1 accuracy and EER (Equal Error Rate) as our main evaluation metrics. 
The ROC (Receiver Operating Characteristic) curves are also plotted for precise comparison
which is based on GAR (Genuine Acceptance Rate) and FAR (False Acceptance Rate).
Refer to~\cite{zhu2020boosting} for their detailed definition.

\subsubsection{Evaluation Protocols} 
To provide comprehensive evaluation and comparison of palmprint verification,
we adopt the matching scheme used in~\cite{Liang2019A,matkowski2019palmprint,Genovese2019palmnet}.
This protocol considers the real application of palmprint verification,
that several images are registered as enrollment and the remaining test images are matched to these images.
Following~\cite{Genovese2019palmnet},
for each palm we take the first 3 images acquired in stage one (see Section~\ref{data_acquisition}) for enrollment and used the remaining images for testing. 
The matching is performed between a test image and the 3 corresponding registered images, 
where the minimum distance of four distances is used as matching distance.

For the learning based methods,
we train the model on the training set and conduct evaluation on the test set.
Non-learning based methods are evaluated only on the test set for a fair comparison.

\subsection{Implementation} 
Our experiments are conducted on a server with 4 Nvidia GTX2080Ti GPUs, 
an Intel Xeon CPU and 128G RAM.  
The proposed method is implemented by PyTorch~\cite{pytorch2019} and code is public available \footnote{\url{https://github.com/dxbdxx/3DCPN}}.
In the first layer,
the length of the Gabor filter is set to 35 and the number of directions is 12.
We find the performances are similar when we change the number of directions.
The candidate sets of hyperparameters are chosen as $\Lambda = \{5, 10, 15\}$ and $\Sigma = \{1, 3, 5\}$ which are commonly used Gabor filters in other algorithms.
The dimension of the palmprint descriptor is 1024.
Before training,
all the layers are initialized by a Gaussian distribution of mean value 0 and standard deviation 0.01.
No pre-training is applied in our method.
We use the stochastic gradient descent (SGD) algorithm with momentum 5e-4 to optimize the total loss. 
The batch size for the mini-batch is set to 64. 
The initial learning rate for the CNN is 1e-2, 
which is annealed down to 1e-4 following a cosine schedule without restart~\cite{loshchilov2017}. 
The total training epochs are 150.

For CUHKSZ, TongJi, and IITD datasets, 
we use official ROI images as input. 
In CUHKSZ dataset,
the reported performances values are cross-validated using 5-fold cross validation.
All the images are normalized and resized to $128 \times 128$. 
No data augmentation is leveraged in our experiments.

\subsection{Main Results}
\subsubsection{Comparison with State-of-the-art Methods}

For palmprint verification, 
we choose coding based methods CompCode~\cite{Kong2004Competitive}, OrdinalCode~\cite{sun2005ordinal}, LLDP~\cite{luo2016local}, CR-CompCode~\cite{zhang2017towards} 
as well as deep learning methods PalmNet~\cite{Genovese2019palmnet}, BPFNet~\cite{li2021BPF}  for our comparison experiments.
The subscript in LLDP method corresponds to the strategy used.
All the experimental parameter settings are the same as reported in the original papers.
The original PalmNet applies 5-fold cross-validation for evaluation and some images are dropped from the datasets.
In our comparison we fix the test set in TongJi and IITD and take all images into account.
We also include several baseline neural networks (Resnet~\cite{he2016deep}, VGG~\cite{simonyan2014very}, GoogLeNet~\cite{szegedy2015going}) and report their performances.
Based on our observation, 
the models pretrained on ImageNet~\cite{deng2009imagenet} could obtain higher performance and thus we only show the results with pretraining. 
For a fair comparison, 
these baseline methods are trained with arc-margin loss and the hyperparameters in the networks are carefully tuned.  
 
The results are shown in Table \ref{tab_test2register} and the best values for each metric are marked in bold.
We can see that our method achieves the best performances on CUHKSZ and TongJi datasets,
compared to both coding based methods and deep learning methods.
However,
the performances of our method on IITD are not satisfactory.
We attribute the relatively poor performances to the small training data size of IITD,
which is a common weakness of deep learning methods.
It should to pointed out that our method still outperforms other methods on IITD.
The results show the effectiveness of 3DCPN across datasets of which the palmprint images are captured in different environments.

\subsubsection{ROC Curves}
The performance gaps among different methods are more distinctive on CUHKSZ dataset as it is larger than most existing datasets.
Here for better visualization, 
we plot the corresponding ROC curves on CUHKSZ dataset in Fig. \ref{fig_roc}.
From the figure we can further demonstrate that our method outperforms other comparison methods.

\subsubsection{Evaluation on Low FARs}
The GAR is the fraction of the genuine scores exceeding the threshold value. 
Considering the high-security requirement of biometric identification,
in real application,
The GARs on low FARs could better exhibit the classification efficiency.
The evaluation results are shown in Table \ref{tab_gar} and the best performances are marked in bold. 
We can see that the proposed approach achieves high GAR while the FAR is low.
For 3DCPN,
the thresholds corresponding to the FARs in the table are 0.17, 0.28, 0.39, and 0.45.

\begin{table}
\caption{GARs @ low FARs (\%) on CUHKSZ dataset}
\begin{center}
	\scalebox{0.95}[1]{
	\begin{tabular}{lcccc}
  	\toprule
	Methods 	& FAR=$10^{-1}$ & FAR=$10^{-2}$ & FAR=$10^{-3}$ & FAR=$10^{-4}$	\\
	\toprule
  	CompCode 	& 99.92			& 99.64         & 99.43 	    & 98.94 \\
  	OrdinalCode	& 99.93  		& 99.29         & 98.94 	    & 97.88 \\
  	LLDP$_1$	& 99.86         & 99.44         & 99.08 	    & 97.95 \\
  	LLDP$_2$	& 99.79         & 99.36         & 97.95 	    & 95.76	\\
  	LLDP$_3$	& \textbf{100}  & 99.79         & 99.51         & 98.94	\\
  	CR-CompCode & 99.71	        & 99.44         & 99.15 	    & 98.73 \\
  	\hline
  	Resnet18 	& 99.85			& 99.22         & 98.23 	    & 96.33 \\
	VGG11-bn 	& 99.43         & 95.34         & 86.30 	    & 70.76	\\
	GoogLeNet 	& 99.58         & 94.21         & 80.64 	    & 58.90	\\
	PalmNet 	& 99.86         & 99.36         & 97.32 	    & 91.24 \\
	BPFNet	    & \textbf{100}  & 99.85			& 99.65			& 98.72 \\
	3DCPN	    & \textbf{100}  & \textbf{100}	& \textbf{99.86}& \textbf{99.44} \\
	\bottomrule
	\end{tabular}}
\end{center}
\label{tab_gar}
\end{table}

\subsubsection{Distance Distribution}

\begin{figure}[t]
	\centering
	\includegraphics[width=\linewidth]{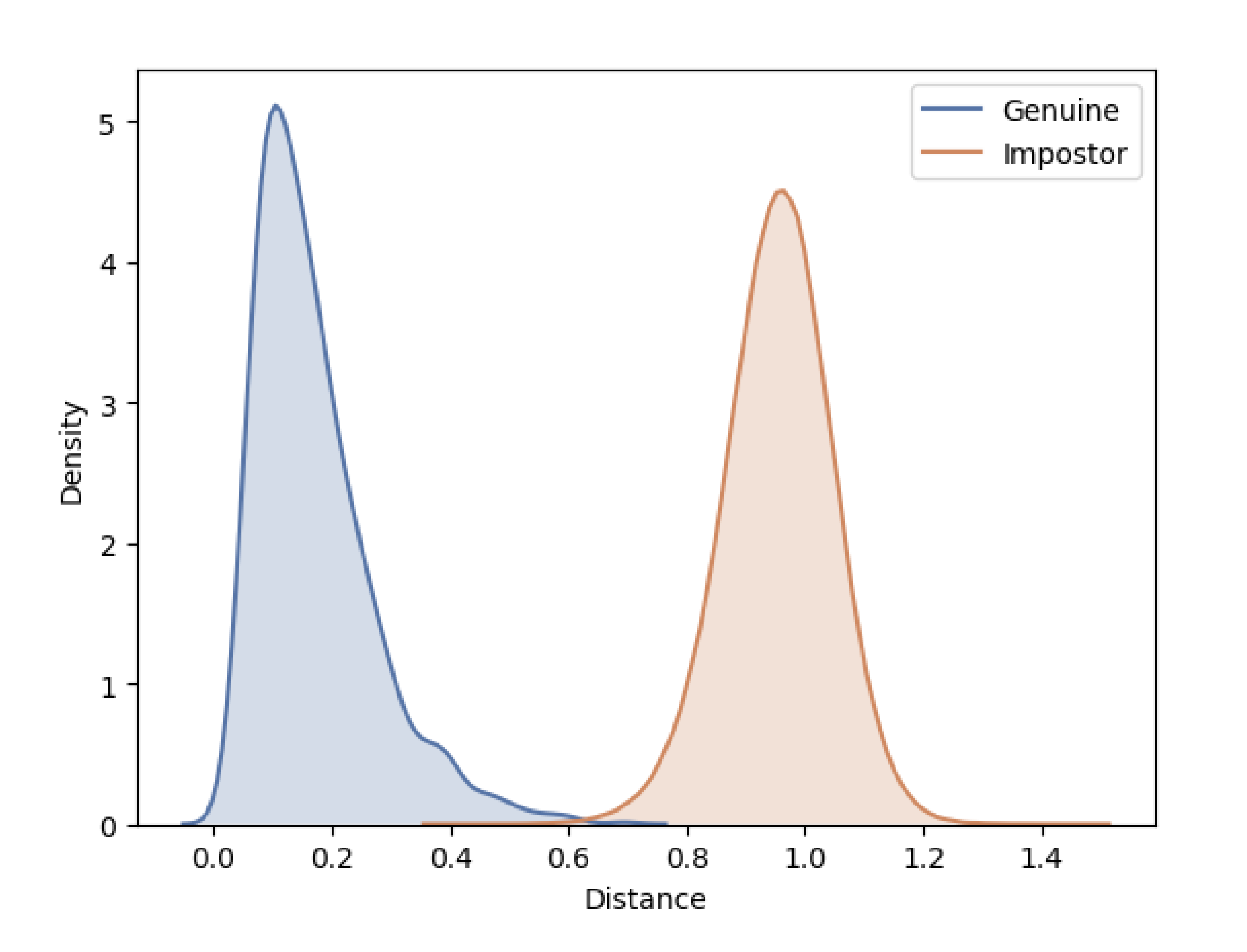}
	\caption{The distance distribution of sample pairs.
	The distribution is calculated by kernel density estimation and is normalized for better visualization.}
\label{fig_dist}  
\end{figure} 

To visualize the distance distribution of learned deep learning model,
we show the final distribution of all possible pairs in Fig. \ref{fig_dist}.
Since we adopt cosine distance in the matching, 
the range of the distance distribution is from 0 to 2.
In the figure,
the distance of total matching pairs is plotted by kernel density estimation and is normalized for better visualization.

\subsection{Robustness to ROI bias}
\label{sec_bias}
\begin{figure}[t]
	\centering
	\includegraphics[width=\linewidth]{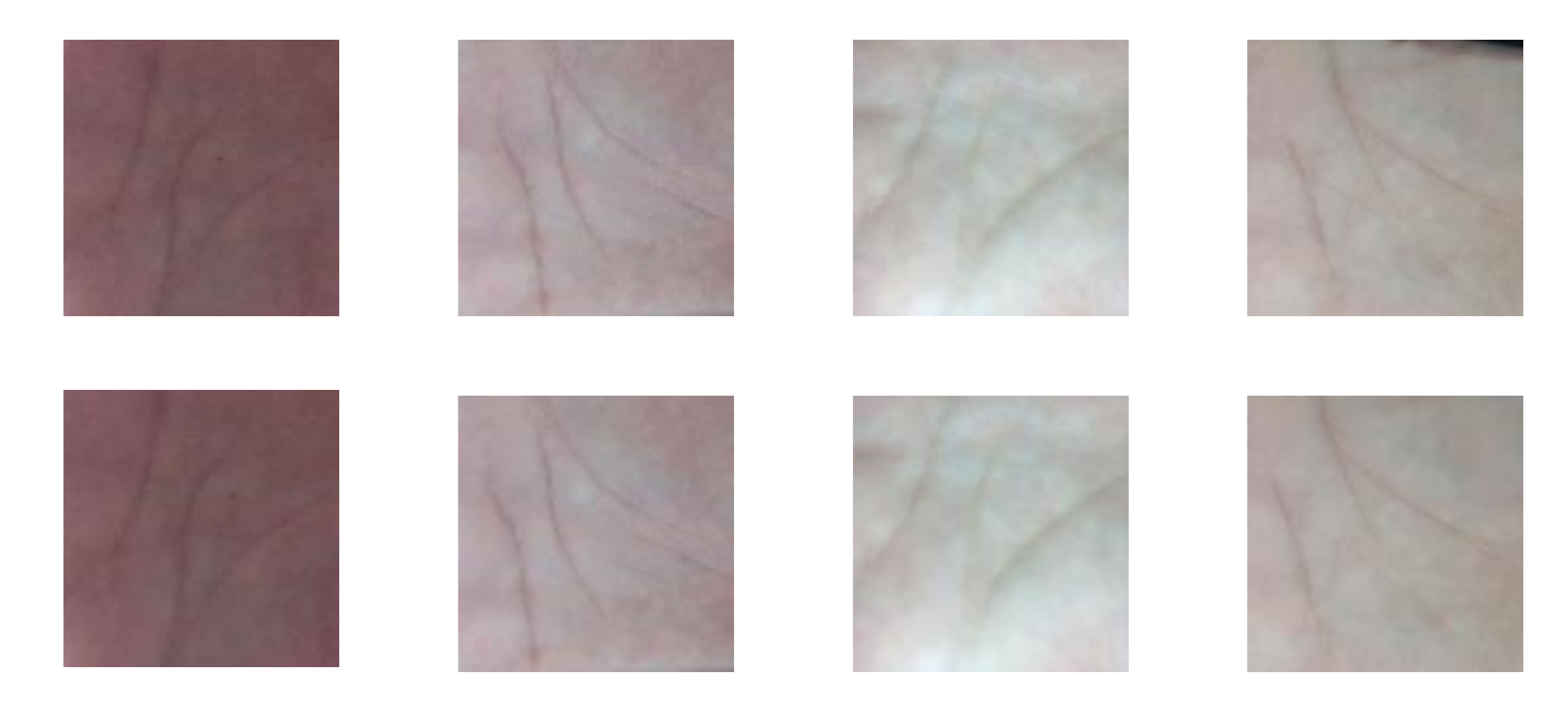}
	\caption{Examples of processed ROIs.
	The first row is the ground truth ROIs and the second row shows the randomly biased ROIs.}
\label{fig_example}  
\end{figure}

3DCPN is a block feature fusion method,
which naturally has the robustness to ROI bias in palmprint images.
Base on our observation,
the majority of palmprint images in CUHKSZ dataset are well aligned.
For simulating the possible ROI bias in the localization process,
we translate the ROIs by random pixels in two directions and resize them to the original size.
Concretely, 
for each image in the test set,
we generate a random translation value from $[r-2, r+2]$ and apply the transformation on the image.
Here, 
$r$ denotes the degree of bias and the larger value of $r$ is more harmful to the final matching.
Some examples are shown in Fig. \ref{fig_example}.

\begin{figure}[t]
	\centering
	\includegraphics[width=\linewidth]{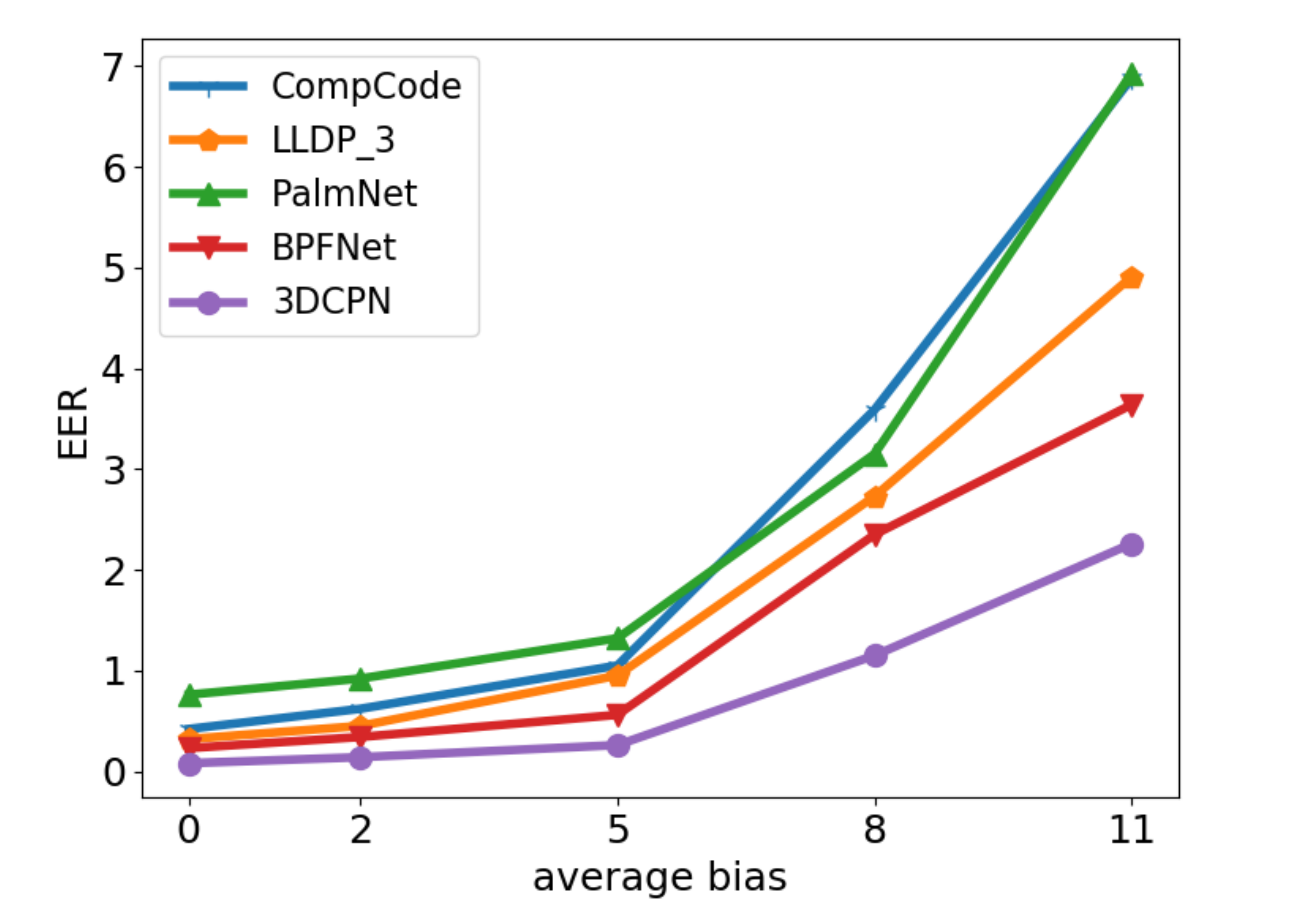}
	\caption{Performance of diffrent methods under ROI bias on CUHKSZ dataset.
	The x-axis is the averaged translation pixels in two direnctions and y-axis shows the EER rates.}
\label{fig_bias} 
\end{figure}

We plot the performance variances in Fig. \ref{fig_bias} of CompCode, LLDP, PalmNet, and 3DCPN.
From the trend we can observe that the performance of the region-matching based method (LLDP$_3$) varies less than that of the pixel-matching based method (CompCode).
Among the methods, 
our 3DCPN is the most stable under ROI bias.
It should be noted that no training image is biased, 
which means the deep learning model is trained with clean data and tested on biased data,
for a fair comparison with coding based methods.

\subsection{Ablation Study}
\subsubsection{Complementary Effect of Curved Gabor Filter}
\label{sec_exp_gabor}

\begin{table}
\caption{Rank-1 accuracy (\%) of palmprint verification methods with three kinds of filters}
\begin{center}
	\scalebox{1}[1]{
	\begin{tabular}{|l||c|c|c|}
  \specialrule{1pt}{0pt}{0pt}
	Filter 			    & CompCode  & LLDP$_3$ 	& 3DCPN	\\
	\hline
	Gabor			    & 99.72		& 99.79		& 97.86 \\
	Curved Gabor	  	& 98.72		& 98.73		& 97.13	\\
	Combined Gabor		& 99.83		& 99.81		& 100 \\
	\specialrule{1pt}{0pt}{0pt}
\end{tabular}}
\end{center}
\label{tab_cgabor}
\end{table}

To demonstrate the complementary effect of the novel proposed curved Gabor filter in line feature extraction,
we design ablation experiments for two representative coding based methods and our 3DCPN,
as shown in Table \ref{tab_cgabor}.
For each method, 
we test the traditional Gabor filter or the curved Gabor filter solely and the Rank-1 accuracy is reported in the first two rows.
The results show that the traditional Gabor filter has a more stable performance than the curved version.
While, as shown in the third row, 
a simple combination strategy can further improve the performance.
In 3DCPN,
the combination is naturally double the filters in the first layer.
And for CompCode and LLDP,
the combined Gabor filter means we average the two distance matrices obtained by each kind of filter.
The results show that the curved Gabor filter can provide additional information on palmprint matching.

\begin{table}[t]
\caption{The performance comparison (\%) of each block and different fusion strategies on CUHKSZ dataset}
\begin{center}
	\scalebox{1}[1]{
	\begin{tabular}{lccc}
	\toprule
	Block / Strategy	&& Rank-1 	& EER		\\
	\toprule
	\# 1 	    		&& 99.72 	& 0.53 	\\
	\# 2 	    		&& 99.42 	& 0.47 	\\
	\# 3 	    		&& 99.13	& 0.62	\\
	\# 4 	    		&& 99.65	& 0.47	\\  
	\# 5 	    		&& 99.79	& 0.41	\\
	\# 6	    		&& 99.09	& 0.55	\\
	\# 7 	    		&& 99.42	& 0.49	\\
	\# 8 	    		&& 99.31	& 0.51	\\
	\# 9 	    		&& 99.04	& 0.92	\\
	\hline
	w/o block loss 		&& 99.67 	& 0.32	\\
	Average 			&& 99.62 	& 0.41	\\
	Max 				&& 99.72 	& 0.36	\\
	Dynamic fusion 		&& 100 	  	& 0.08	\\
	\bottomrule
	\end{tabular}}
\end{center}
\label{tab_block}
\end{table}

\begin{table}
\caption{Influence of hyperparameters in arc-margin loss function}
\begin{center}
  \scalebox{1}[1]{
  \begin{tabular}{|l|cc|cc|}
  \specialrule{1pt}{0pt}{0pt}
  Loss function	& s 	& m		& Rank-1 	& EER	\\
  \specialrule{1pt}{0pt}{0pt}
  softmax Loss	& - 	& -		& 99.23 	& 0.47 \\
  \hline
  arc-margin Loss	& 64 	& 0.5	& 99.86 	& 0.19 	\\
  arc-margin Loss	& 32 	& 0.5	& 99.72 	& 0.25 	\\
  arc-margin Loss	& 16 	& 0.5	& 100		& 0.08 	\\
  arc-margin Loss	& 64 	& 0.3	& 99.83 	& 0.32 	\\ 
  arc-margin Loss	& 64 	& 0.7	& 99.79 	& 0.32 	\\
  \specialrule{1pt}{0pt}{0pt}
  \end{tabular}}
\end{center}
\label{tab_arc}
\end{table} 

\subsubsection{Effectiveness of Dynamic Fusion}
To show the effectiveness of the dynamic fusion, 
we provide the performances of each block feature in our final trained model in Table \ref{tab_block}.
The final learned weights (before softmax) for each block is $W=\{-0.8, 2.1, -0.7, 3.1, 3.0, 0.5, -0.2, 0.8, -2.5\}$.
The performance gap among blocks in fact shows their importance to the final feature map.
For instance, 
the main-lines of a palm is generally not in the last block,
which causes the corresponding feature not distinctive compared to other regions.
We also set three basic fusion methods,
which are trained from scratch,
as comparison baselines.
The first method abandons the "feature ensemble" part in 3DCPN and the supervision is the single arc-margin loss on the top.
The second method simply averages all the block features in the fusion part and other training settings are the same as our model. 
The last method is similar to the second method where the average operation is replaced by max-pooling.
From the table we can see that though each single block feature is not discriminative enough, 
our fusion strategy outperforms other methods by fusing the block dynamically.

\subsection{Influence of Hyperparameters}
\subsubsection{Influence of Arc-margin Loss Function}
\label{sec_arc}
There are two hyperparameters in the arc-margin loss function,
the scale multiplication $s$ and the target angular margin $m$.
To investigate their influences on the performance,
we conduct experiments with some common settings.
The experiment results are shown in Table \ref{tab_arc}.
The result shows that, 
compared to softmax loss,
arc-margin loss can boost the performance in palmprint recognition task,
bringing at least $0.11\%$ performance gain on EER rate.
The performance varies less when we change $s$ compared to change $m$,
which means the model is more sensitive to the parameter $m$.
The best performance is achieved when $s=16$ and $m=0.5$.

\subsubsection{Sensitiveness of Trade-off Parameter $\mu$}
\label{sec_sensitiveness}
\begin{figure}[t]
  \centering
  \includegraphics[width=\linewidth]{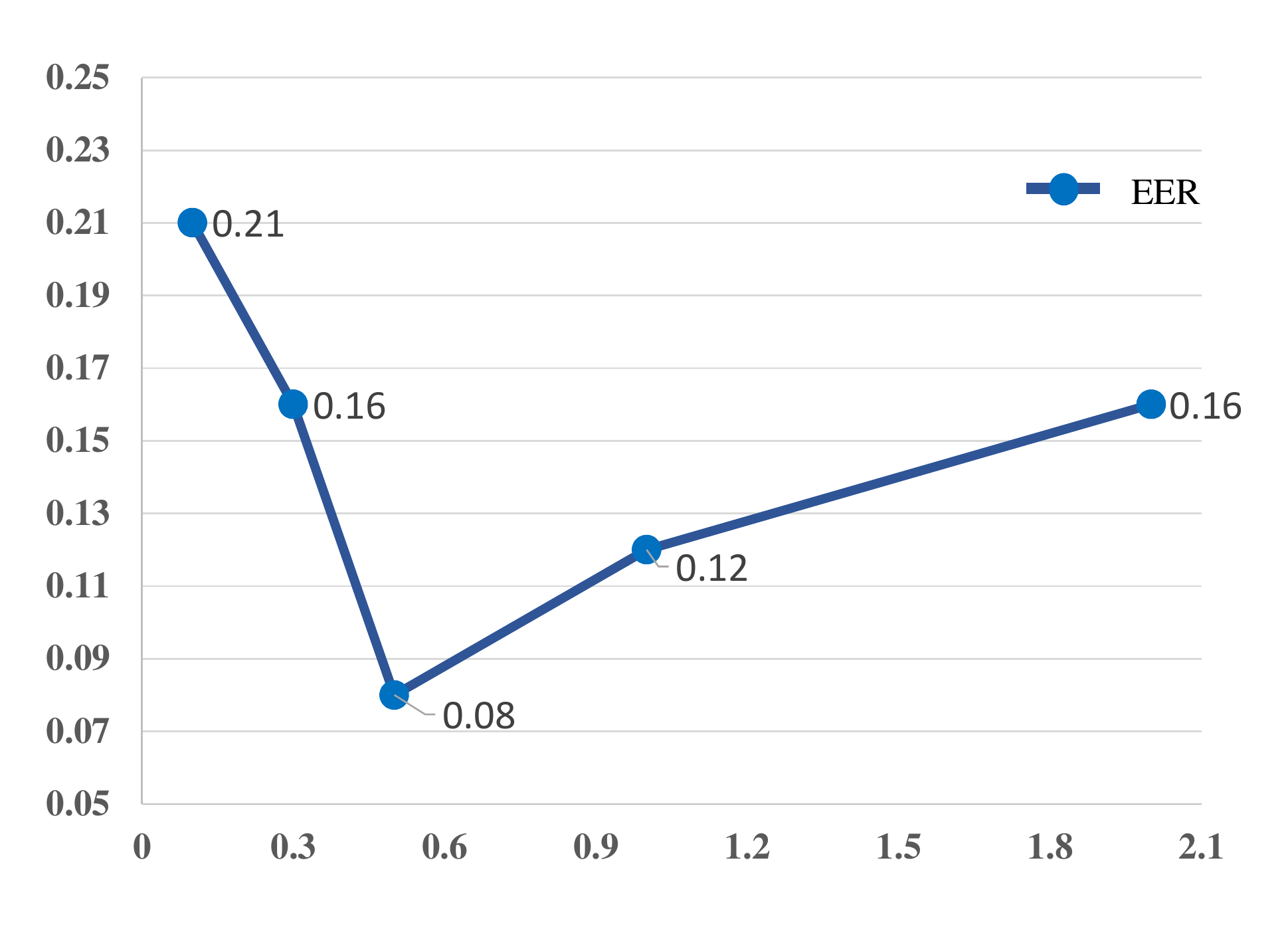}
  \caption{The recognition performance with different loss weights on CUHKSZ dataset.}
\label{fig_mu} 
\end{figure}

To observe the effect of loss weight $\mu$ on the performance, 
we conduct experiments supervised by arc-margin loss on CUHKSZ dataset. 
The experiment results are shown in Fig. \ref{fig_mu}.
A larger $\mu$ could enhance the block features while a too large value would make the training hard,
which is detrimental to the final result.
In the five experiments,
the loss weight $\mu$ is set to 0.1, 0.3, 0.5, 1, and 2 respectively.
We can see that the EER rate first decreases and then augments.
3DCPN can obtain the best result when the loss weight is $0.5$.

\section{Conclusion}
\label{sec_conclusion}
In this paper,
we build a large-scale touchless palmprint dataset CUHKSZ-v1 and propose a novel deep learning based framework called 3DCPN.
The framework is designed for large-scale palmprint recognition,
which leverages Gabor features and 3D convolution layers.
In 3DCPN, 
we embed two kinds of Gabor filters for low-level feature extraction and reform it to have 3D shapes.
For the enhancement of local palmprint features,
we employ a block loss to supervise the network training. 
Finally we conduct comprehensive experiments including convincing ablation studies to demonstrate the efficiency of our framework.
In the future, 
we will develop palm detection algorithms and tackle the ROI misalignment problem during ROI localization.

\section*{Acknowledgement}
This work is supported by Shenzhen Institute of Artificial Intelligence and Robotics for Society, 
and Shenzhen Research Institute of Big Data. 
The work is also supported by the NSFC fund (61906162),
Open Project Fund from Shenzhen Institute of Artificial Intelligence and Robotics for Society (AC01202005017), 
and China Postdoctoral Science Foundation (2019TQ0316, 2019M662198, 2020TQ0319, 2020M682034).
The authors also would like to thank all the volunteers who contributed their palm images for this
study.

\bibliography{references}{}
\bibliographystyle{IEEEtran}

\ifCLASSOPTIONcaptionsoff
  \newpage
\fi

\end{document}